\documentclass{article}

\usepackage{microtype}
\usepackage{graphicx}
\usepackage{subfigure}
\usepackage{booktabs} 

\usepackage{hyperref}

\usepackage[accepted]{icml2019}

\icmltitlerunning{Iterative Model-Based RL Using the DNC}

\begin{document}

\twocolumn[
\icmltitle{Iterative Model-Based Reinforcement Learning Using Simulations in the Differentiable Neural Computer}



\icmlsetsymbol{equal}{*}

\begin{icmlauthorlist}
\icmlauthor{Adeel Mufti}{fiveai,uoe}
\icmlauthor{Svetlin Penkov}{fiveai,uoe}
\icmlauthor{Subramanian Ramamoorthy}{fiveai,uoe}
\end{icmlauthorlist}

\icmlaffiliation{fiveai}{FiveAI, Edinburgh, United Kingdom}
\icmlaffiliation{uoe}{The University of Edinburgh, Edinburgh, United Kingdom}

\icmlcorrespondingauthor{Adeel Mufti}{adeelmufti@gmail.com}

\icmlkeywords{Reinforcement Learning, Differentiable Neural Computer, Lifelong Learning, Model-Based Reinforcement Learning, Simulations}

\vskip 0.3in
]



\printAffiliationsAndNotice{}  

\begin{abstract}
We propose a lifelong learning architecture, the Neural Computer Agent (NCA), where a Reinforcement Learning agent is paired with a predictive model of the environment learned by a Differentiable Neural Computer (DNC). The agent and DNC model are trained in conjunction iteratively. The agent improves its policy in simulations generated by the DNC model and rolls out the policy to the live environment, collecting experiences in new portions or tasks of the environment for further learning. Experiments in two synthetic environments show that DNC models can continually learn from pixels alone to simulate new tasks as they are encountered by the agent, while the agents can be successfully trained to solve the tasks using Proximal Policy Optimization entirely in simulations.
\end{abstract}

\section{Introduction}
\label{intro}
The Differentiable Neural Computer (DNC) is an architecture where external memory is attached to an Artificial Neural Network (ANN), akin to conventional computers where random-access memory is attached to a central processing unit \cite{graves2016hybrid}. The external memory gives the DNC additional capacity, as the memory provides the ANN with the capability to track and manipulate the input data directly rather than through the ANN weights and neuron activity. The authors of the DNC show that it can learn to solve complex algorithmic tasks that are difficult for traditional ANNs -- from sorting lists to graph traversal \cite{DBLP:journals/corr/GravesWD14, graves2016hybrid}. The authors train the DNC using curriculum learning across all tasks, and observe by inspecting its memory contents that it learns and improves algorithms over the curriculum steps.

AI agents can be trained to perform tasks in simulations generated by a predictive model of the environment \cite{sutton1991dyna,gu2016continuous,ha2018recurrent}. Where such a model is learned using ANNs, we hypothesize that DNCs present a robust mechanism that is well suited for lifelong learning. The algorithmic capabilities of the DNC can be leveraged to simulate multiple tasks present in an environment, which can be useful for continual model-based Reinforcement Learning (RL). To test this, we devise an architecture for lifelong RL where an ANN agent and DNC model are trained in conjunction to iteratively learn to solve tasks in a given environment. We call our architecture the Neural Computer Agent (NCA). In the NCA, we use a Convolutional Autoencoder to learn low dimensional latent state representations from the frames, consisting of pixels, collected by the agent in the environment. The DNC model is trained to predict the environment states in this latent space, and the agent is trained to use the latent states and information from the model. We use the model to simulate the environment entirely in the latent space, where the agent is trained using RL. The agent then rolls out to the live environment to collect new experience which is used to further train the DNC model.

We first set out to test the algorithmic and lifelong learning capabilities of the DNC in isolation, on an addition task. We train the DNC with curriculum learning where the task progressively gets more complex at each curriculum step, and find that the DNC indeed learns to infer an algorithm and generalizes earlier than traditional ANNs by an order of magnitude. We then test the NCA on two synthetic toy environments: a partially observable directional path navigation environment, and a fully observable obstacle-based grid navigation environment. In both environments, the DNC model successfully learns new tasks while retaining older ones, and the accompanying RL agent successfully learns to solve the tasks in the live environment when trained using Proximal Policy Optimization (PPO) in simulations generated by the DNC model alone.

\section{Related Work}
While the authors of the DNC conduct an experiment to train it using RL on a block puzzle task, they use low dimensional hand-coded states on a single task \cite{graves2016hybrid}. Our approach learns states from pixels, and the DNC is used to learn a predictive model of dynamic environments with multiple tasks present.

The idea of training an RL agent using simulations is present in prior work. The Dyna Architecture uses an action model to train an agent in ``hypothetical" states \citet{sutton1991dyna}. Normalized advantage functions (NAF) uses iteratively refitted local linear models for training agents in ``imaginations" using Q-learning \cite{gu2016continuous}. World Models is a model-based policy learning architecture, where (a) a Convolutional Variational Autoencoder (CVAE) is used to learn a latent state representation, (b) a Mixture Density Network output from a Long Short Term Memory (LSTM) network is used to learn a predictive model of the environment, and (c) Evolution Strategies are used to learn a policy in a simple controller using ``dreams" \cite{ha2018recurrent}. The NCA is closely related to World Models but differs in that we use a simpler Convolutional Autoencoder (CAE) for state representation learning, the predictive model is learned by a DNC instead of LSTM, the predicted states are output directly instead of being sampled through a Gaussian Mixture Model, and the agent is a feedforward ANN trained using PPO.

The goal of lifelong learning in AI is to be able to retain knowledge when trained on sequentially new tasks, and selectively transfer prior knowledge to learning the new tasks \cite{silver2013lifelong}. For ANNs, methods for lifelong learning when encountering new tasks vary between constraining network weights using regularization, dynamically expanding capacity of the network while freezing the old capacity, or keeping a buffer of data from previous tasks to mix into data from the new tasks \cite{DBLP:journals/corr/abs-1805-12369}. The NCA keeps a buffer of data from previous experience for training the DNC model. In the context of RL, notable work on lifelong learning includes gradient-based meta-learning algorithms for fast and efficient adaption to new tasks \cite{DBLP:journals/corr/abs-1710-03641}, maintaining a mixture of models that are instantiated and recalled as new or old tasks are encountered \cite{DBLP:journals/corr/abs-1812-07671}, or learning local models for new tasks that can be bootstrapped with a globally maintained model \cite{DBLP:journals/corr/abs-1803-11347}. Our approach is closest to \citet{DBLP:journals/corr/abs-1803-11347} in learning a global model, but when encountering new tasks we leave the adaptation entirely unsupervised to the DNC, allowing it to leverage its algorithmic capabilities to transfer knowledge between multiple tasks.

State Representation Learning involves learning low dimensional latent representations from the high dimensional states an agent observes. The representation captures the variation in the environment based on the agent’s actions. This representation helps overcome the curse of dimensionality and improves speed and performance of policy learning algorithms such as RL. Our method of learning a latent state is similar to others where a Convolutional Autoencoder is used to learn a compressed latent representation from pixels observed by an agent \cite{munk2016learning,ha2018recurrent,DBLP:journals/corr/abs-1802-04181}.

\section{The Neural Computer Agent}
\label{sec:arch}
At the core of our architecture, the Neural Computer Agent (NCA), is a DNC that is leveraged to learn a robust model of the environment. The DNC contains an inherent algorithmic bias as it is made to use its external memory to track relevant information, and its external memory also provides it with additional capacity unavailable to traditional ANNs. Its induced algorithms can quickly adjust to multiple new tasks, which we show in Section~\ref{ssec:exp-addition}. Thus, we aim to learn a global model of multiple tasks in a particular environment, in a single DNC-based model, as opposed to using individual task-specific models. We allow the DNC to continually adapt to new tasks based solely on the experience collected by the agent.

\begin{figure}[ht]
\vskip 0.2in
\begin{center}
\centerline{\includegraphics[width=\columnwidth]{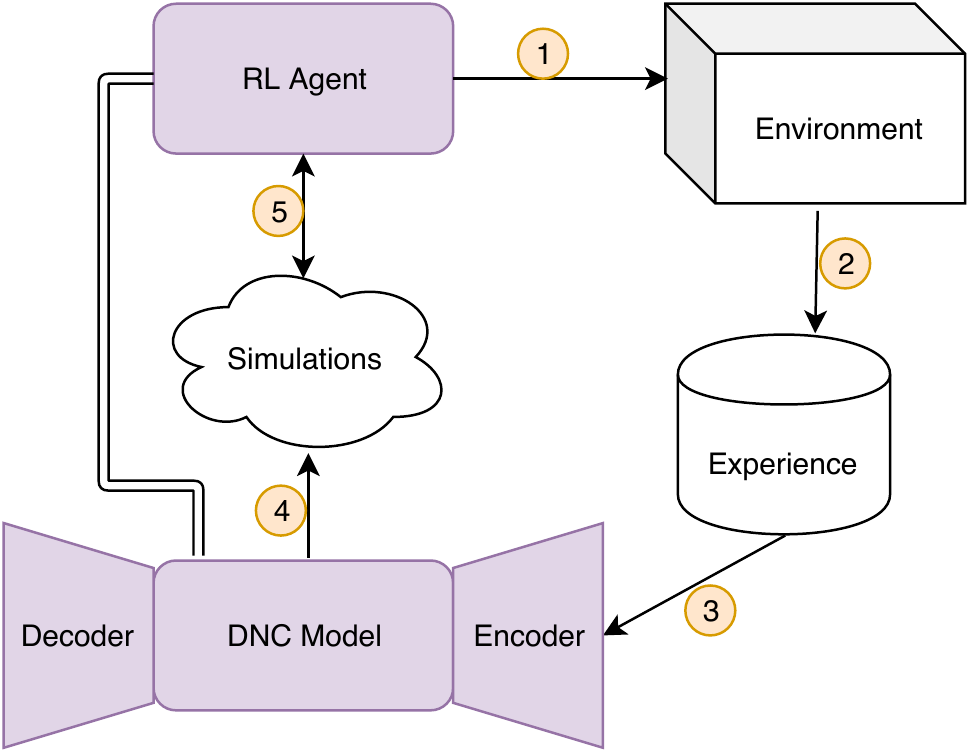}}
\caption{Schematic for our iterative lifelong RL architecture, the {\bfseries Neural Computer Agent}. An agent interacts with the environment to collect experience, which is used to train a predictive DNC model. The model is then used to simulate the environment to train the agent. The agent then rolls out to the environment again to collect new experience and iterate again.}
\label{fig:mtrl-dnc-architecture-schematic}
\end{center}
\vskip -0.2in
\end{figure}

The schematic for the NCA is shown in Figure~\ref{fig:mtrl-dnc-architecture-schematic}. Each iteration consists of the AI agent performing rollouts in the live environment to collect experience. The experience is then used to train the DNC model to predict the next states, rewards, and probabilities that a rollout is done. The model is then used to generate simulations to train the RL agent to learn a policy, after which the agent is tested in the live environment. At that point an iteration is complete, and the next one starts where the agent rolls out again to the environment, to collect more experience using the policy it has learned in the previous iteration. This process continues until satisfactory results are achieved, such as an average cumulative reward in an environment. The process is detailed in Algorithm~\ref{alg:architecture}.

\begin{figure}[ht]
\vskip 0.2in
\begin{center}
\centerline{\includegraphics[width=\columnwidth]{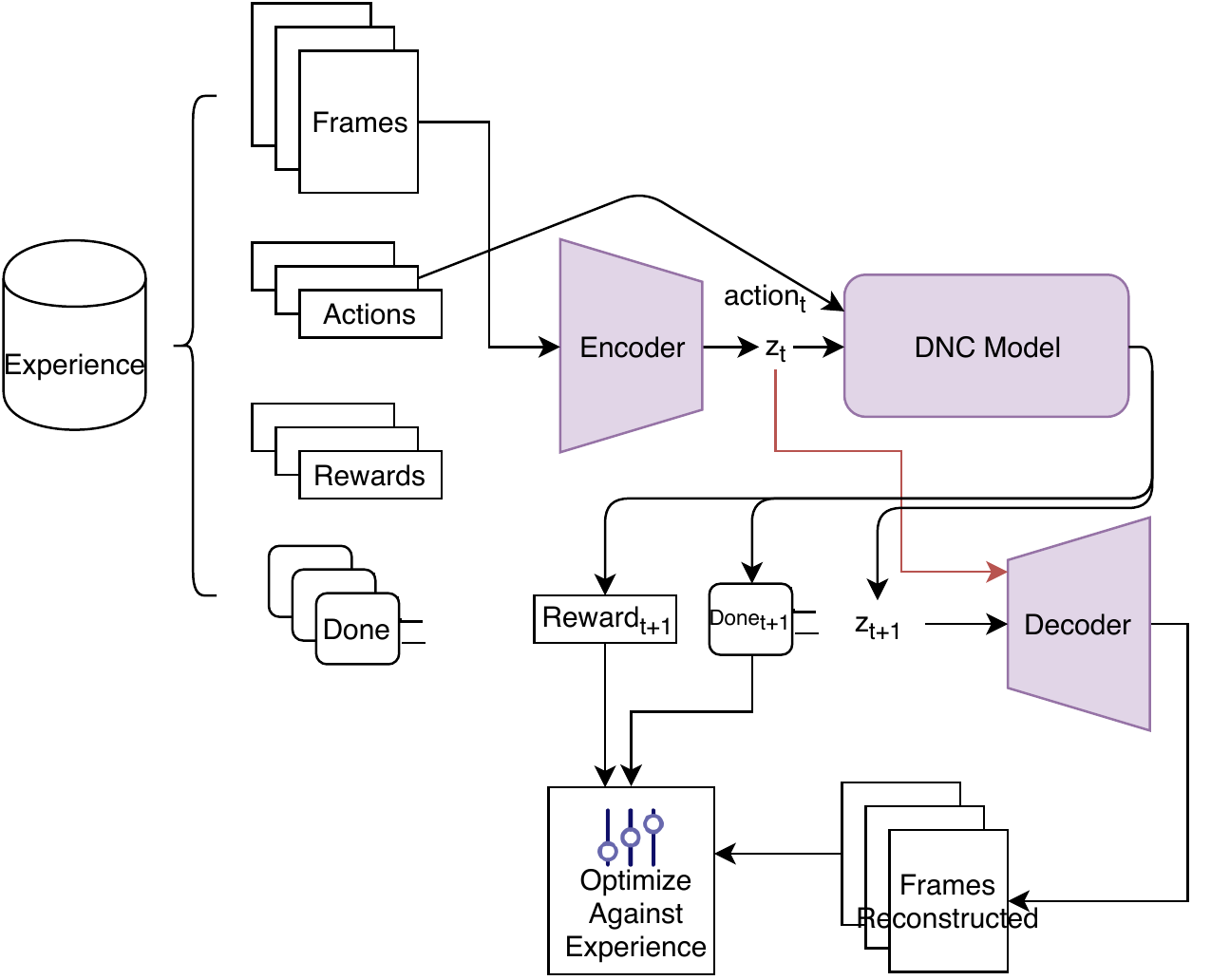}}
\caption{The DNC model architecture and training procedure. The model is trained using experience collected by the agent in the live environment as ground truths. Frames at timestep $t$ are encoded into latent states $z_t$ which are concatenated with $action_t$ to produce predictions $z_{t+1}$, $reward_{t+1}$, $done_{t+1}$.}
\label{fig:mtrl-dnc-model-schematic}
\end{center}
\vskip -0.2in
\end{figure}

Details for our model are illustrated in Figure~\ref{fig:mtrl-dnc-model-schematic}. The model is composed of the DNC, which is coupled with a CAE. The CAE learns a latent state representation, $z$, for observations from the live environment. The task of the DNC at each timestep $t$ is to take as input the current state $z_t$ and an action $action_t$ to predict at the next time step $t+1$ the next state $z_{t+1}$, reward $reward_{t+1}$ and probability that the rollout is done $done_{t+1}$. We train the DNC and the CAE encoder and decoder end-to-end, to optimize a loss function consisting of the following:

\begin{enumerate}
	\item Negative log-likelihood of a Bernoulli distribution for the CAE
	\item Cross entropy for the predicted done probability
	\item Squared error for the predicted reward
	\item Squared error for the predicted latent state
\end{enumerate}

During early experiments we observe that by converting pixels to grayscale, normalizing between 0 and 1, and treating them as being drawn from a Bernoulli distribution, works best when computing the negative log-likelihood between pixels from the ground truth frames and those reconstructed from the latent state space by the CAE's decoder. For our simulations, we need a probability at each timestep that the rollout is done, thus cross entropy between the done prediction and the true/false ground truths is most appropriate. The reward is necessary for training an agent using RL in simulations, so we simply regress the value by taking the squared error between the predicted rewards and ground truth rewards at each timestep. The CAE learns a direct unconstrained latent state representation (as opposed to with CVAEs where Gaussian parameters regularized to the unit normal distribution are learned to sample the latent state), so at each timestep we take the squared error from the predicted latent state for the next timestep and the ground truth latent state for the next timestep. Note that to obtain latent state ground truth $z_{t+1}$ during training, the current approximation is used from the CAE. Though this could potentially lead to convergence problems, in practice this works adequately.

\begin{algorithm}[h]
   \caption{The Neural Computer Agent}
   \label{alg:architecture}
\begin{algorithmic}[1]
   \STATE {\bfseries initialize:} $iteration$=0, DNC model weights $\theta_{m}$, convolutional encoder/decoder weights $\theta_{ed}$, agent weights $\theta_{a}$, experience buffers $\zeta_m$ and $\zeta_a$
   \REPEAT
       \STATE {\bfseries collect} in experience buffer $\zeta_m$ 99\% of $N_m$ rollouts using $\theta_a$ from live environment
       \STATE {\bfseries optimize} $\theta_{m}$, $\theta_{ed}$ end to end for $E_m$ epochs over $\zeta_m$; discard experience in $\zeta_m$ except 1\% of $N_m$ rollouts
       \REPEAT
           \STATE {\bfseries collect} in experience buffer $\zeta_a$, $N_a$ simulated rollouts using $\theta_{m}$, $\theta_{ed}$, $\theta_a$ 
           \STATE {\bfseries optimize} $\theta_{a}$ for $E_a$ epochs over $\zeta_a$; discard experience in $\zeta_a$
           \STATE {\bfseries rollout} agent using $\theta_{m}$, $\theta_{ed}$, $\theta_a$ in live environment for $N_t$ rollouts and measure average cumulative reward
       \UNTIL{either 0:$E_{RL}$ cycles reached or current task in curriculum solved based on average cumulative rewards}
       \STATE $iteration$+=1
   \UNTIL{all tasks solved}
\end{algorithmic}
\end{algorithm}

\begin{figure}[ht]
\vskip 0.2in
\begin{center}
\centerline{\includegraphics[width=\columnwidth]{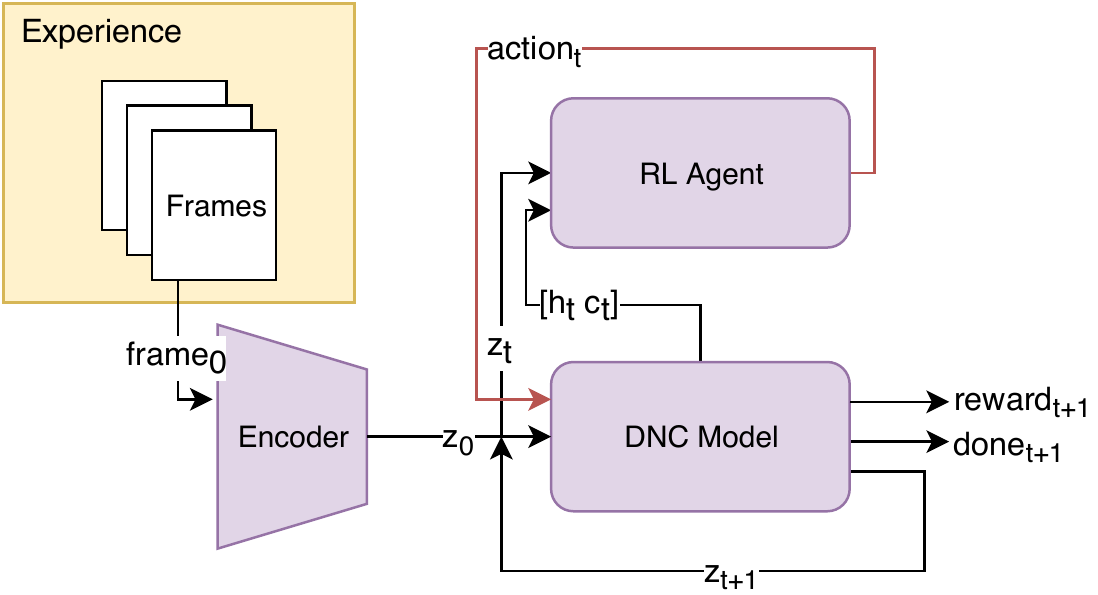}}
\caption{Illustration for how a simulation is generated using the model, and how an agent receives inputs from the model.}
\label{fig:mtrl-dnc-simulation}
\end{center}
\vskip -0.2in
\end{figure}

How a simulation is generated using the DNC model is illustrated in Figure~\ref{fig:mtrl-dnc-simulation}. To seed the simulation at timestep $t=0$, a frame randomly sampled from live rollouts from experience buffer $\zeta_m$ is used after being encoded to $z_0$, though in our case initial frames per environment are all the same. Then, concatenated $[z_0 + action_0]$ are input to the model which produces predictions $z_{1}$, $reward_{1}$, and probability $done_{1}$ at timestep $t+1$. After which, each $z_{t+1}$ state prediction and action produced by the agent are input back into the model, to produce the next prediction, until $done_{t+1}$ reaches a fixed threshold probability.

The agent is composed of a feedforward ANN and is trained in local ``RL cycles" within each iteration, where the agent rolls out and collects experience in simulations generated by the model, optimizes over the experience using PPO, and tests the learned policy in the live environment to log performance with cumulative rewards achieved. Depending on what is desired of the architecture, there can be a fixed $E_{RL}$ cycles performed at each iteration where $E_{RL}\ge1$, or until a desired average cumulative reward is achieved. At each timestep $t$, the agent receives as input the concatenation of the latent state vector $z_t$, and hidden and cell state vectors $h_t$, $c_t$ from the LSTM controller in the DNC. This is similar to how the Controller receives input from the CVAE and LSTM Model in the World Models framework \cite{ha2018recurrent}, and works well in practice. 

\section{Experiments}
\subsection{Setup}
For all our experiments, we keep the DNC architecture similar to the ones used in experiments of \citet{graves2016hybrid}. We use a LSTM controller with 256 units, an external memory size of length 256, width 64, and 4 read heads, and a batch size of 1. We use the same CAE used by \citet{mnih2015human}, and give our latent state representation (the $z$ vector) 32 dimensions. For our RL experiments, the output of the DNC is of 34 dimensions, where 32 dimensions are for the next predicted latent state representation, 1 for the predicted reward, and 1 for the predicted probability that a rollout is done. We use the Adam optimizer with a learning rate of $\alpha=1e-4$. The model is trained with Truncated Backpropagation Through Time using a sequence length of 64.

The hyperparameters and settings for Algorithm~\ref{alg:architecture} are as follow. Experience buffer $\zeta_m$ has the capacity to hold $N_m=1000$ rollouts from the live environment, where the frames observed, actions performed, and rewards received per rollout are all stored. At each iteration, we keep 1\% of rollouts from the previous iteration in $\zeta_m$, to expose the agent to previous data. The frames observed are converted to grayscale, resized to 64px x 64px, and pixels normalized between 0 and 1. The model weights $\theta_m$, $\theta_{ed}$ are optimized for $E_m=10$ epochs. Experience buffer $\zeta_a$ consists of $N_a=128$ simulated rollouts, until $done_{t+1} >= 0.75$. Agent weights $\theta_a$ are optimized for $E_a=5$ epochs. The agent is tested in the live environment over $N_t=100$ rollouts.

For the agent we use an ANN with a single hidden layer of 256 units. The agent's weights $\theta_a$ are optimized using simulated experience collected in $\zeta_a$, where we use the PPO algorithm using hyperparameters $\epsilon=0.1$, $\gamma=0.99$ and $\lambda=0.95$. We use the Adam optimizer with a learning rate of $\alpha=1e-4$. 

\subsection{DNC capabilities}
\label{ssec:exp-addition}
We design a toy addition task to test the lifelong learning and algorithm inferring capabilities of the DNC. The task involves receiving a sequence of one-hot encoded integers as input and producing their sum as output. In a single input sequence, integers are randomly selected between 0 and 2 inclusive initially, while the length of the input sequence (total number of integers being summed) is randomly selected between 1 and 12 inclusive initially. We then progressively introduce higher value integers up to 9 inclusive, and progressively increase the length of the sequence up to 100. As the full input is fed to the DNC sequentially one encoded integer at a time, we add a delimiter field to the input vector where 0 signifies more integers will be input and 1 signifies the end of the input sequence. Thus the input vectors $x$ is 11 dimensional, and the output $y$ is a single number representing the sum. We train the DNC using Mean Squared Error between predicted sum $\hat{y}$ and the ground truth sum $y$. Additionally, we train a LSTM network alongside to compare performance with, where the network consists of 256 units to match the capacity of the LSTM controller in the DNC.

\begin{figure}[ht]
\vskip 0.2in
\begin{center}
\centerline{\includegraphics[width=\columnwidth]{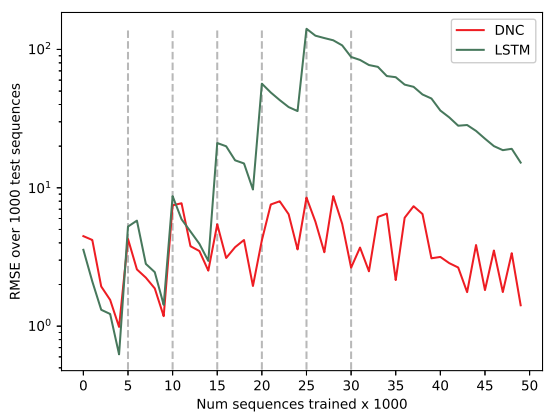}}
\caption{DNC and LSTM performance over the course of training on multiple progressively difficult addition tasks. The curriculum steps switch at every $5,000$ sequences trained, until $30,000$ sequences, where the most difficult task remains fixed until training concludes.}
\label{fig:dnc-lstm-addition-performance}
\end{center}
\vskip -0.2in
\end{figure}

Curriculum training on this task involves 6 steps, designed to present progressively more difficult tasks that clearly build on prior knowledge:
\begin{enumerate}
	\item Sum 1 to 12 integers of values between 0 and 2
	\item Sum 1 to 12 integers of values between 0 and 5
	\item Sum 1 to 12 integers of values between 0 and 9
	\item Sum 1 to 25 integers of values between 0 and 9
	\item Sum 1 to 50 integers of values between 0 and 9
	\item Sum 1 to 100 integers of values between 0 and 9
\end{enumerate}
Each curriculum step is trained for $5,000$ total sequences, and tests of the current curriculum step are performed over $1,000$ sequences throughout training. The tests consist of $1,000$ randomly generated sequences where the Root Mean Squared Error (RMSE) between ground truth sum $y$ and predicted sum $\hat{y}$ are logged. A graph of these test results that depict the learning progress of the DNC and LSTM over the course of training is shown in Figure~\ref{fig:dnc-lstm-addition-performance}. Note that at the end of curriculum step 6 ($30,000$ sequences onward), the task remains the same until training is concluded at $50,000$ sequences.

The results show the DNC clearly outperforming in this lifelong learning task. It can be observed that there is a peak in RMSE for both architectures as curriculum steps change. The RMSE over the test sequences for the DNC is slightly unstable compared to LSTM, which we attribute to a lack of hyperparameter tuning such as learning rate. While both LSTM and DNC struggle initially, the DNC is able to leverage knowledge from prior tasks as the addition problem gets more difficult through the curriculum steps, and performs significantly better than LSTM until LSTM catches up slowly. We thus conclude that the DNC presents a promising framework in the context of multitask and lifelong learning.

\subsection{Directional Path Navigation}
\label{ssec:directional-path-navigation}
The first environment we construct to test the NCA consists of a straight path navigation task where the path is represented as numbered tiles in ascending order starting from 0. The tiles increment by 1 and the path ends at the goal tiles -- 9, 19 or 29, depending on which level the agent is on. At each timestep, the agent can perform one of two action: move up the path where the numbers increment, or move down the path where the numbers decrement. The path is partially observable, where the current frame (observation) shows which numbered tile the agent is on. The agent always starts at tile 0, and the objective of the environment is to reach the goal tiles in a minimum number of steps, to unlock the next level of the path. On level 1 of the environment, the agent will need to reach tile 9 in 9 steps by continuously moving up the path, otherwise will not be able to move past tile 9. Once the agent solves level 1, level 2 is unlocked which consists of tiles 10 through 19 where the same rule applies -- reach goal tile 19 in a minimum number of steps. And the same rules apply for level 3, from tiles 20 through 29, where the path and environment end at tile 29.

\begin{figure}[h!]
\centering
\subfigure[Level 1]{\label{fig:directional-path-navigation-curriculum-1}\includegraphics[width=\columnwidth]{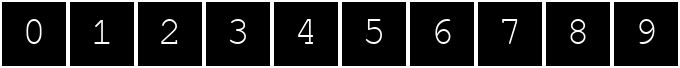}}\\
\subfigure[Level 2]{\label{fig:directional-path-navigation-curriculum-2}\includegraphics[width=\columnwidth]{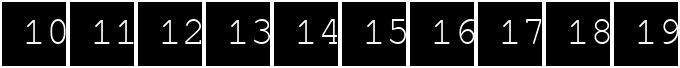}}\\
\subfigure[Level 3]{\label{fig:directional-path-navigation-curriculum-3}\includegraphics[width=\columnwidth]{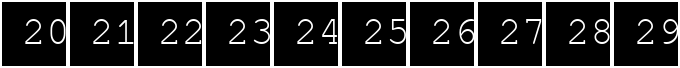}}\\
\caption{The numbered tiles composing a path at each level in the Directional Path Navigation environment. The agent observes a tile at a time and has to learn to traverse up the full path in a minimum number of steps.}
\label{fig:directional-path-navigation-curriculum}
\end{figure}

The reward given at each timestep is -1 unless the agent lands on the goal tile, where the reward is +0.1 if the agent does not reach it in a minimum number of steps, and +10 if it does reach it in a minimum number of steps. If the ultimate goal tile at level 3, tile 29, is not reached in 50 total timesteps, the rollout ends. Levels 1 and 2 are considered solved when average cumulative scores of +2 and +3, respectively, are achieved over 100 rollouts in the live environment. The environment is considered solved when an average cumulative score of +4 is achieved over 100 rollouts in the live environment.

\begin{figure}[ht]
\vskip 0.2in
\begin{center}
\centerline{\includegraphics[width=\columnwidth]{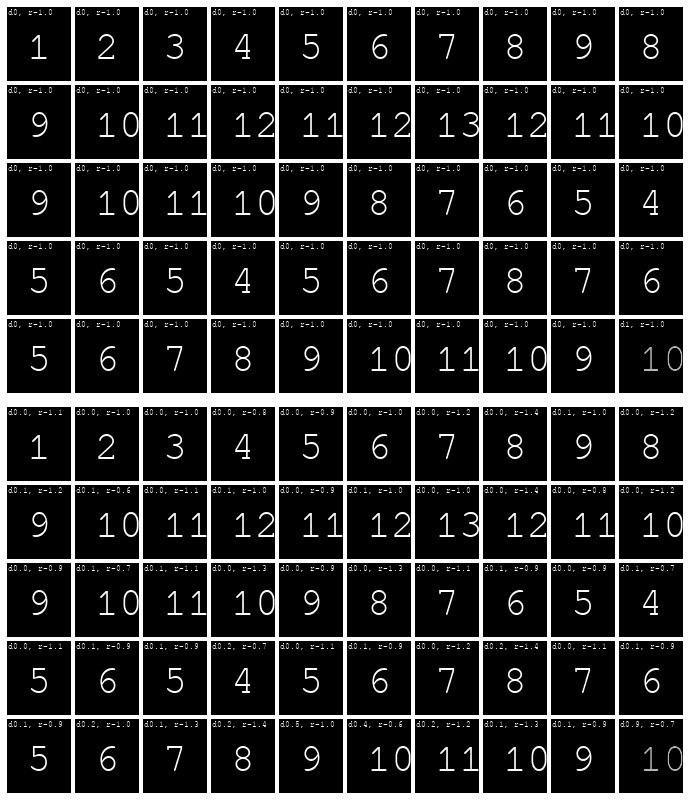}}
\caption{A rollout in the {\bfseries Directional Path Navigation} environment. Depicted across each row are the individual frames (each numbered tile) sequentially observed by the agent at each timestep. The top collage contains frames from the live environment where the agent performs a series of actions. The bottom collage contains reconstructed frames from a simulation generated from the DNC model using the same series of actions. The labels above each frame indicate the rewards and done probabilities. It can be seen that the DNC model adequately learns to simulate the environment.}
\label{fig:directional-path-navigation-rollouts}
\end{center}
\vskip -0.2in
\end{figure}

As the core focus of this work is on learning a model of an environment in the DNC in the multitask and lifelong learning context, this environment is simplistic by design and should be easy to solve for any RL agent. The simple discrete states and objectives makes it easy to measure the predictive capabilities of the model when rolled forward in simulations. 

We perform 3 total iterations using the NCA on this environment, to match the 3 levels (tasks) present in the environment. For each iteration, we enforce a curriculum by locking the live environment from progressing to the next level prematurely, in order to investigate and measure the performance of the model and agent in isolation on a level at a time.

We observe that the model successfully learns to reproduce the environment in simulations. This includes predicting the next state, predicting the next reward, and predicting whether the rollout has concluded (the done probability). Figure~\ref{fig:directional-path-navigation-rollouts} shows frames from a rollout where the agent collects experience in the live environment, which is then simulated using the model with the same actions. The predicted frames and done probability appear to be correct in all cases checked, though the reward seems off in some cases but points in the correct general direction. We also note that the agent successfully learns to solve the curriculum task at each iteration, in just one RL cycle. Over 100 test rollouts in the live environment, the agent achieves an average cumulative score of +2.0$\pm$0.0, +3$\pm$0.0, and +4$\pm$0.0 on levels 1, 2, and 3 respectively.

\subsection{Obstacle-Based Grid Navigation}
We design a second environment to test the NCA's lifelong learning capabilities over multiple tasks, where prior knowledge can be leveraged. The environment consists of a 5 x 5 grid at each level, where an agent has to learn to navigate to the goal cell in a minimum number of steps, while avoiding obstacles. At each timestep, the agent can perform one of four navigation actions: up, down, left, and right. The agent always starts in the top left cell of the grid, and the goal is always at the bottom right of the grid.

\begin{figure}[h!]
\centering
\subfigure[Level 1]{\label{fig:obstacle-grid-navigation-curriculum-1}\includegraphics[width=64px]{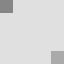}}
\subfigure[Level 2]{\label{fig:obstacle-grid-navigation-curriculum-2}\includegraphics[width=64px]{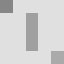}}
\subfigure[Level 3]{\label{fig:obstacle-grid-navigation-curriculum-3}\includegraphics[width=64px]{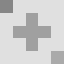}}
\caption{The levels (tasks) present in the Obstacle-Based Grid Navigation environment. The levels are designed to progressively increase in difficulty by adding obstacles. The agent starts at the top left cell of the grid, and has to reach the goal cell at the bottom right in a minimum number of steps.}
\label{fig:obstacle-grid-navigation-curriculum}
\end{figure}

In this environment, we present 3 progressively challenging levels (tasks). The first level contains no obstacles, and the agent simply has to learn to get to the goal cell in a minimum number of steps to progress to the next level. The second level contains a vertical contiguous 3-cell obstacle in the middle of the grid, where the agent has to navigate around the obstacle to get to the goal cell in a minimum number of steps. The third and final level builds on the second level and has an additional horizontal contiguous 3-cell obstacle in the middle of the grid. The levels are shown in Figure~\ref{fig:obstacle-grid-navigation-curriculum}.

At each timestep the agent receives a reward of -0.1 when it moves onto an empty cell that is not the goal cell, -1 when it tries to navigate out of bounds or through an obstacle, +0.1 when it arrives on the goal state but not in the minimum possible number of steps, and +10 when it arrives on the goal cell in the minimum possible number of steps, after which it progresses to the next level. The rollout ends at a maximum of 50 timesteps, unless the agent arrives at the ultimate goal state in the third level in the minimum possible number of steps. Level 1 is considered solved if the agent is able to achieve an average cumulative reward of +9.3 over 100 rollouts in the live environment, +18.6 on level 2, and +27.9 on level 3.

To experiment on this environment using the NCA, we again perform 3 total iterations, to match the 3 levels present in the environment. For each iteration, we again enforce a curriculum by locking the live environment up to the level number corresponding to the current iteration number, to prevent the agent from progressing to the next level prematurely, in order to isolate the tasks being learned. This assists in measuring and investigating NCA's performance in a step-wise manner.

\begin{figure}[h]
\vskip 0.2in
\begin{center}
\centerline{\includegraphics[width=\columnwidth]{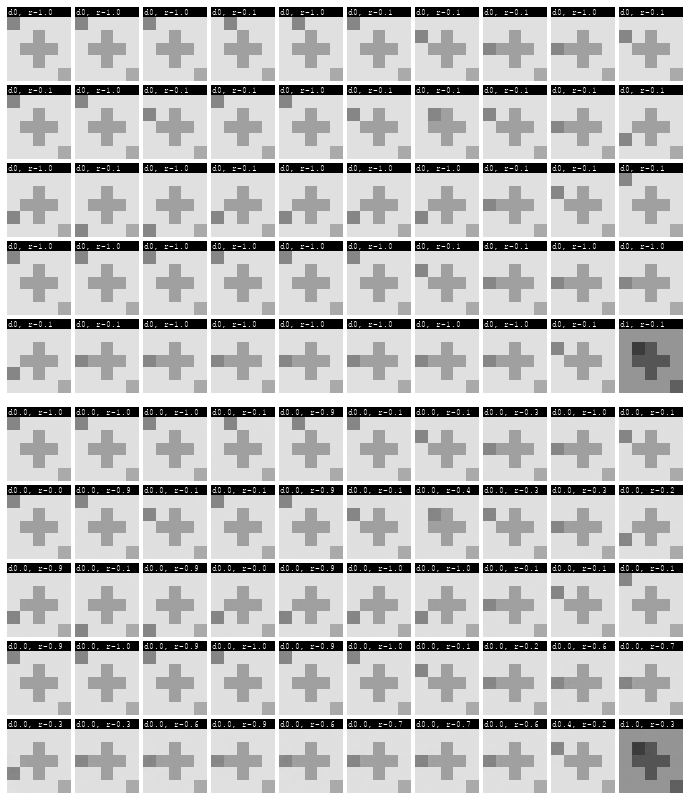}}
\caption{A rollout in the {\bfseries Obstacle-Based Grid Navigation} environment. The individual frames sequentially observed at each timestep are depicted across the rows. On the top are frames from the live environment where the agent performs a series of actions. On the bottom are reconstructed frames from a simulation using a learned DNC model performed with the same actions. The labels above each frame indicate the rewards and done probabilities.}
\label{fig:obstacle-grid-navigation-rollouts}
\end{center}
\vskip -0.2in
\end{figure}

Figure~\ref{fig:obstacle-grid-navigation-rollouts} shows frames from a rollout where the agent collects experience in the live environment, which is then simulated using the model with the same actions performed. We observe that the predicted frames and done probabilities appear correct in all simulations examined, though again the reward is observed to be slightly off but pointing in the correct general direction.

At each iteration, the agent learns to solve the level at the current curriculum step. Over 100 test rollouts in the live environment, the agent achieves an average cumulative score of +9.3$\pm$0.0, +18.6$\pm$0.0, and +27.9$\pm$0.0 on levels 1, 2, and 3 as they are unlocked at each curriculum step. As opposed to the experiments on the directional path navigation environment, the agent takes more than one RL cycle to converge to a solution at each iteration, which we attribute to the environment being more difficult and due to the agent having to select from more possible actions.

\section{Discussion And Future Work}
Our main focus is on the capabilities of the DNC incrementally learning a predictive model of environments that involve multiple progressive tasks. Thus, through the course of the experiments, we observe the predictions generated by the DNC model in simulations. Since the states are represented as a latent vector, we use the decoder to reconstruct the predicted states into frames (images) for the purpose of visual inspection. We can then take as ground truths frames from a rollout from the live environment where a particular sequence of actions are performed, and simulate the environment in the DNC model with the same sequence of actions, to visually inspect the predicted frames against the ground truths. As our environments are not stochastic, we should expect to see the exact frames in predictions as the ground truths. We visually inspect the reconstructed predictions through the iterations of training and observe adequate results, though perhaps a more quantitative and comprehensive evaluation is necessary, such as measuring accuracy through a classifier for the possible states. For the rewards and done probabilities, we inspect and find in all cases checked that though the predicted rewards are not exact, they correlate with the ground truths, and the done probability only increases as a task is solved or maximum number of steps are reached. Here again a quantitative measure of prediction accuracy is warranted in future work.

For the directional path navigation environment, the environment is highly simplistic, and it may be that the agent learns to entirely ignore all inputs and always produce the same action (go up the path), which would solve the environment. While this is potentially an issue for validating the iterative architecture as a whole, we focus on the predictive capabilities of the DNC in learning a model of an environment incrementally, which we observe as described above. In future work, the directional path navigation task could be extended in ways to require the agent to make use of both actions.

The environments tested are simple for the purpose of establishing proof of concept in using a global DNC model in the continual RL setting. The environments represent multiple tasks requiring an RL agent and predictive model to learn continually. In the directional path navigation environment, the numbered tiles change for the same navigation task as the agent solves each level. In the obstacle-based grid navigation environment, new obstacles are presented, requiring the agent to adapt its navigation strategy. Other work on continual learning in the RL context examines environments where the tasks involve locomotion of agents represented as half-cheetahs, ants, or spiders \cite{DBLP:journals/corr/abs-1803-11347,DBLP:journals/corr/abs-1710-03641,DBLP:journals/corr/abs-1812-07671}. The tasks are adapted in a number of ways such as varying the nonstationary locomotion environment the agent is in (the background, the slope of the terrain, and so on), or crippling the agent by disabling a joint or a leg. The agent's continual learning capabilities are tested as the tasks are varied. An extension of our work would be to test the NCA on such environments.

Another area to explore involves the particulars of the DNC's external memory. It could be useful and interesting to analyzing the memory contents through the timesteps of simulations, to get a grasp on its inner workings as it moves through different parts (tasks) of an environment. It is also not certain what an adequate memory size is. Whereas we use the memory length, width and read heads settings used by \citet{graves2016hybrid} universally, either smaller or larger memory sizes may be more optimal depending on the type of environment and tasks. This memory could also be exposed to the RL agent to provide it with potentially useful information the DNC is tracking while producing predictions.

\section{Conclusion}
We present an iterative model-based RL architecture, the Neural Computer Agent (NCA), geared towards lifelong learning through simulations of an environment. At the core of the architecture is a predictive model learned in a DNC. The DNC is chosen as it can infer algorithms and performs well in multitask and curriculum learning settings, as shown by its authors and tested by our own toy addition task. We test the NCA on two synthetic environments where multiple tasks are unlocked progressively as the agent learns to solve levels of the environment over the iterations. In each environment, we observe an adequately learned representation of the environment by the DNC model by inspecting how it performs in predicting next states, rewards, and probabilities of a rollout being done. The agent in the NCA is successfully able to solve each level in the environments tested by iteratively training entirely in simulations generated by the DNC model, and rolling out to the live environment for testing.

\bibliography{refs}
\bibliographystyle{icml2019}

\end{document}